\documentclass[letterpaper]{article}
\usepackage[preprint]{aaai2027}
\usepackage[hyphens]{url}
\usepackage{graphicx}
\usepackage{natbib}
\setcitestyle{aysep={,}}
\usepackage{bibentry}
\usepackage{caption}
\usepackage{tabularx}
\usepackage{array}
\usepackage{makecell}
\usepackage{booktabs}
\title{CDEP Agent: Connecting Meteorologically Detected Temporal Compound Events to Real-World Documentary Evidence}
\author{
Zhuoran Li\textsuperscript{\rm 1}\equalcontrib,
Weiyi Kong\textsuperscript{\rm 2}\equalcontrib,
Boer Zhang\textsuperscript{\rm 3}
}
\affiliations{
\textsuperscript{\rm 1}The University of Hong Kong\\
\textsuperscript{\rm 2}University of Toronto\\
\textsuperscript{\rm 3}Harvard University
}

\begin{document}

\maketitle
\shortcites{jones2022,vicenteserrano2010,stein2020}

\begin{abstract}
Compound drought-to-extreme-precipitation (CDEP) events are recognized in climate science as a growing driver of extreme impact, but whether this recognition carries over into real-world early warning and post-event documentation is unknown, so a meteorologically real CDEP event may pass with neither advance warning nor any later record. Here we present CDEP Agent, an auditable LLM-agent framework that tests this mismatch directly by linking CDEP candidates detected from meteorological reanalysis to real-world hazard and impact evidence across sources with different spatial scales, temporal resolutions, and reporting conventions. Using California as a case study, we identify 408 candidate CDEP events from ERA5 observations during 2021–2025 and evaluate each against the U.S. Drought Monitor, NOAA Storm Events, and public webpages along five dimensions: antecedent drought, extreme rainfall, local impact, hazard-impact attribution, and explicit drought-to-rainfall linkage. Only 34.3\% of candidates are corroborated on both hazard components, and just 1.5\% are ever explicitly linked to their antecedent drought, indicating that most meteorologically detected CDEP events go undocumented and their compound nature almost never enters the record at all. Our framework gives climate scientists a way to test physical event definitions against what actually gets documented, and gives social scientists, economists, and disaster-response agencies a provenance-linked evidence base for compound events that current warning and reporting systems largely fail to capture.
\end{abstract}

\section{Introduction}

Climate change is intensifying hydroclimatic extremes worldwide, and a comprehensive understanding of their real-world consequences is essential for early warning, adaptation planning, and disaster risk management. A growing body of work points to drought and extreme-precipitation events as among the clearest signatures of this shift \citep{rodell2023}, events that are not only becoming more frequent but also more severe \citep{gu2023,satoh2022}. What is more striking is how quickly the climate system can swing between these two extremes: rapid transitions from drought to intense rainfall have grown markedly more common worldwide since 1980, a pattern that points to these transitions increasingly behaving as compound events in their own right, rather than as two unrelated hazards that happen to occur close together \citep{qing2023}. The Intergovernmental Panel on Climate Change (IPCC)'s Sixth
Assessment Report identifies compound extremes as an important
category of climate risk, while recent events illustrate their severe
impacts across regions \citep{ipcc2021,zscheischler2025}. Yet a compound event being meteorologically real is not the same as it being documented. A drought-to-extreme-precipitation transition can be clearly present in reanalysis data and still leave little trace in the records that governments, journalists, and disaster databases actually produce, because those records are written around discrete storms,
discrete droughts, and discrete administrative jurisdictions,
not around the meteorologically defined sequence connecting them
\citep{jones2022,li2026wikimpacts}. This is true even though compound events have by now been carefully formalized as a distinct class of hazard in their own right, with typologies and conceptual frameworks describing how they arise from the joint or sequential occurrence of drivers that individually might not be extreme \citep{leonard2014,zscheischler2020}. Even recent efforts to make disaster archives more usable, such as GDIS \citep{rosvold2021} and Geo-Disasters \citep{teber2025}, add spatial precision to existing single-hazard entries without asking whether a compound sequence was recognized as such in the first place. Meanwhile, meteorological detection of compound events has grown increasingly capable \citep{yin2025}, which only sharpens the question: when a detector flags hundreds of candidate compound events from meteorological data, how many of them were ever noticed, reported, or acted on by anyone outside the climate science community, and what determines whether an event clears that bar?

To address this, we present CDEP Agent, an auditable LLM-agent framework that connects meteorologically detected CDEP candidates to the real-world hazard and impact evidence that would indicate they were recognized. Language agents show promise for scientific retrieval and synthesis
\citep{skarlinski2024,asai2026} and domain-specific decision support
\citep{xie2025}, but require rigorous task-level evaluation
\citep{chen2025}. CDEP is one example of the broader class of compound climate events and consists of a prolonged drought period followed rapidly by intense rainfall within a short window. Given a candidate specified by a county, an antecedent drought period, and a subsequent rainfall window, the agent retrieves and aligns evidence from sources such as the U.S. Drought Monitor, NOAA Storm Events, and public webpages, and evaluates each candidate along five explicit dimensions: whether the antecedent drought was recorded, whether the subsequent extreme rainfall was recorded, whether it produced a local impact, whether that impact was attributed to the hazard, and whether any source explicitly links the drought to the later extreme-precipitation event. Each decision retains its source provenance, supporting quotations, and date, location, and same-event matching results, and a conservative labeling protocol ensures that an unresolved or negative outcome is recorded as such rather than treated as evidence that nothing happened.

CDEP serves as a useful demonstration case for several reasons. It is well characterized meteorologically \citep{qing2023,deng2024}, it admits a threshold-based definition over meteorological data, and California in particular has seen a documented rise in this kind of precipitation volatility (\citealp{swain2018}; \citealp{swain2025}). California's own history illustrates how disruptive this kind of whiplash can be. Between 2012 and 2017, the state endured a five-year drought of record severity, only for the drought to end abruptly when an intense storm brought heavy rainfall across the region. The sudden influx of water damaged the spillway at Oroville Dam so severely that roughly 190,000 residents downstream had to be evacuated \citep{vahedifard2017}. Determining whether other CDEP candidates were recognized requires matching federal drought monitors, storm logs, and local news to the candidate's place, time, and hazard. Manual review does not scale to hundreds of candidates or support efficient reruns as records or detection thresholds change.

Applying the CDEP Agent to 408 candidate drought-to-extreme-precipitation transitions identified in California produces a structured, provenance-linked record for each candidate; only 34.3\% are corroborated on both drought and extreme-precipitation components by official or public records, and just 1.5\% are ever explicitly connected to their antecedent drought by any source. This gap has direct consequences for the early-warning, adaptation, and disaster-response decisions that depend on knowing which hazards were actually recognized on the ground. Yet where corroborating webpages exist, they typically capture substantive, real-world consequences, such as flooding, evacuations, power outages, and property damage, showing that the gap lies in whether an event is documented at all, rather than in what can be recovered once it is. We then use these structured, provenance-linked records to ask what separates the candidates that entered the documentary record from those that did not: whether meteorological intensity and persistence or local impact predict documentary recognition, as distinct from cases where the gap simply reflects limitations in evidence availability or retrieval. By making these documentary gaps explicit and auditable, our framework gives climate scientists a systematic way to test and refine meteorological candidate definitions against what actually gets recognized as societally relevant, and gives social scientists and economists a provenance-linked evidence base for studying how and when compound climate events come to be known.

\section{Methods}
Our study consisted of two separate stages: reanalysis-based candidate construction and agent-based evidence review. First, we identified historical CDEP candidates in California from ERA5-Land using a fixed operational definition based on Precipitation--Evapotranspiration Index (SPEI-3), a location-specific extreme-precipitation threshold, and a maximum drought-to-extreme precipitation  interval of three months. This stage was completed before the agent was applied. It produced, for each candidate, a county, an antecedent drought period, a subsequent extreme precipitation event window, and the corresponding candidate characteristics. The agent did not select or modify the meteorological thresholds used to construct these candidates.

\begin{figure*}[t]
    \centering
    \includegraphics[width=0.88\textwidth]{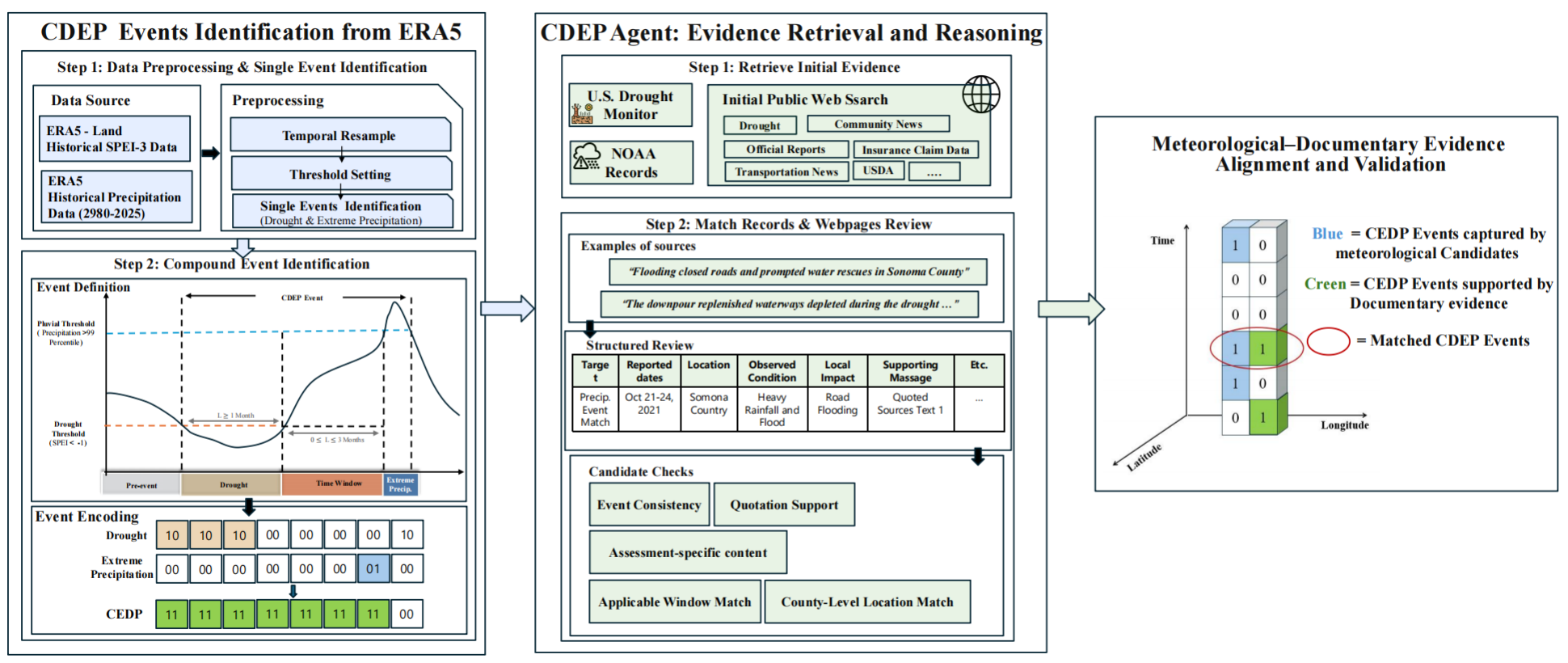}
    \caption{Overview of the CDEP Agent workflow. Meteorological
    candidates are first constructed from ERA5-Land drought and
    extreme-precipitation data. For each fixed candidate, the agent
    retrieves official records and public webpages, structures the
    retrieved evidence, and applies temporal, spatial, same-event,
    and quotation-support checks. The resulting documentary evidence
    is then aligned with the meteorological candidate to produce
    provenance-linked assessments.}
    \label{fig:cdep_pipeline}
\end{figure*}

Given these fixed candidate records, the agent first matched each candidate to structured official records and conducted an initial search of public webpages. It then checked whether qualifying evidence remained missing for the antecedent drought, the subsequent extreme precipitation event, local impacts, or an explicit connection between the drought and the later extreme precipitation event, and directed follow-up searches to the missing items. For each retrieved source, the agent extracted the reported dates, locations, drought or extreme precipitation conditions, documented consequences, stated relationships, and supporting quotations. Deterministic checks then applied the time window and location requirements corresponding to the component under review. For extreme precipitation-event impacts and hazard--impact
connections, the checks also required the hazard and consequence
to refer to the same event. Figure~\ref{fig:cdep_pipeline}
summarizes the complete workflow, from ERA5-Land candidate
construction through evidence retrieval, structured source review,
candidate-specific checks, and meteorological--documentary
alignment.

\subsection{Constructing Historical CDEP Candidates from Meteorological Data}

We characterized antecedent drought using the three-month SPEI-3 derived from ERA5-Land
\citep{munozsabater2019}. SPEI measures the balance between precipitation
and atmospheric water demand over a selected accumulation period; the
three-month scale was used here to represent moisture conditions preceding
the extreme precipitation transition \citep{vicenteserrano2010,vicenteserrano2020}. For candidate construction, a grid cell was classified as being under
drought when SPEI-3 was below $-1$, with more negative values
indicating more severe drought \citep{deng2024}. This SPEI-based rule was used only to construct the meteorological candidate set; the later antecedent-drought assessment used independent documentary evidence from the U.S. Drought Monitor and qualifying public webpages.

Extreme precipitation events were independently detected using daily precipitation data from ERA5-Land \citep{munozsabater2019}, based on three-day accumulated precipitation (PR3). For each grid cell, the P99 threshold was calculated from the 1985--2014 reference period, and precipitation episodes exceeding this threshold were classified as extreme rainfall events. This is consistent with event-based flood classifications that
distinguish short- and long-rainfall processes \citep{stein2020}, and
with flood-risk studies in China that use three-day rainfall maxima as
an indicator \citep{li2012}. Together, these studies support anchoring
the extreme-precipitation leg of the compound event at a high,
flood-relevant percentile rather than a generic extreme-rainfall
definition.

The study focuses on detecting extreme precipitation events within
three months after drought, following prior CDEP work that evaluates
one-, two-, and three-month transition intervals \citep{deng2024}. This enables a more comprehensive identification of CDEP events. 

Together, the SPEI-3 threshold, the location-specific 99th percentile of three-day precipitation, and the maximum three-month drought-to-extreme precipitation interval define the operational CDEP specification used to construct the candidate set in this study. These literature-grounded choices provide one consistent way to identify candidate transitions; they are not intended to represent the only valid definition of CDEP. Because the agent receives candidate locations and event windows as inputs, the same retrieval and evidence-review procedure can be rerun on candidate sets constructed using alternative thresholds, although the present evaluation uses only the specification reported here.
\subsection{Searching Public Records and Webpages}

After the meteorological candidates had been constructed, each fixed candidate record entered the agent-based retrieval and evidence-review stage.
The retrieval stage began by aligning each candidate with structured records from the U.S. Drought Monitor and NOAA Storm Events using the available county, date, event-type, and consequence fields. It also conducted an initial public-web search covering agency webpages, local and regional news, and retrospective reports. Web queries combined the candidate location and relevant dates with terms describing the evidence being sought, such as drought, water shortage, drought-related restrictions, heavy rainfall, flooding, road disruption, property damage, power interruption, evacuation, or agricultural loss. The purpose was to find records describing the specified place and event period, rather than webpages that discussed drought or flooding only in general terms.

After the initial search, the agent reviewed the evidence already collected. Follow-up searches were selected from four evidence targets. An antecedent-drought search was used when no qualifying source documented drought during the candidate drought window. An extreme-precipitation-event search was used when no qualifying source documented the subsequent extreme precipitation event. Impact-focused searches were used when consequences of a matched extreme precipitation event were absent or incompletely described. Once both hazard components were supported, a linkage-focused search could be used when no source explicitly connected the antecedent drought to the later extreme precipitation event. The searches therefore differed across candidates according to the evidence already available.

Follow-up queries were required to preserve the candidate location, the time period relevant to the selected search target, and the type of evidence being sought. Drought-focused queries used the antecedent drought window, whereas extreme precipitation-event and impact-focused queries used the subsequent extreme precipitation-event window. Linkage-focused queries could include both periods. Queries that omitted these elements, repeated an earlier search, remained too broad to identify the candidate event, or introduced unsupported storm names, locations, damage claims, casualties, or other event details were rejected before retrieval. Previously visited webpages were also excluded from later searches. These checks kept each search tied to information already present in the candidate record or in an accepted source.

Search-result snippets and automatically generated summaries were used only to locate possible sources. A government record or webpage entered the evidence review only after the underlying record or page text had been retrieved. The same maximum query and page limits were applied to every candidate so that the search procedure remained comparable across cases. When a potentially relevant source could not be retrieved or reviewed, that limitation was recorded and carried into the final assessment rather than treated as evidence that no event or impact occurred.

\subsection{Reviewing and Structuring Retrieved Sources}

Retrieved sources were reviewed as descriptions of the candidate drought period or extreme precipitation event rather than as collections of matching words. A page mentioning drought or flooding in the candidate county was not sufficient if it referred to conditions outside the applicable candidate window, described a different event or period, or reported only a forecast or warning. Each source therefore had to be matched to the applicable candidate window, county, and hazard before it could support a decision.

Structured government records were read from their fixed fields, including event dates, locations, event type, damage, casualties, and other recorded consequences. Each webpage was processed separately together with the corresponding candidate information. The language model determined whether the page contained no relevant record, one clearly identifiable drought period or extreme precipitation event that could be matched to the candidate, or multiple periods or events that could not be separated reliably. For a clearly identified record, it extracted the reported dates, locations, observed drought or extreme precipitation conditions, local impacts, statements connecting those impacts to the extreme precipitation hazard, any explicit connection between the antecedent drought and the later extreme precipitation event, and the passages supporting those fields.

Every webpage field used as evidence had to be supported by text that could be located in the retrieved page. Information from different storms was not combined, and a hazard reported for one event could not be paired with an impact reported for another. When a page contained several events that could not be separated, it was not used to support a positive decision.

The language model handled the parts of the review that required interpretation of webpage prose, including identifying the event being described, distinguishing event dates from publication dates, locating the event, identifying observed hazards and consequences, and recognizing relationships explicitly stated by the source. Deterministic checks then applied the candidate window corresponding to the component under review. Drought evidence had to overlap the antecedent drought window, whereas extreme precipitation-event evidence had to overlap the subsequent extreme precipitation-event window. In both cases, the source had to identify the candidate county or a clearly located place within it. For local impact and hazard–impact assessments, the documented consequence and extreme precipitation hazard also had to refer to the same matched event.

A publication or update date could not substitute for the date on which the event occurred. Statewide reports, broad regional descriptions, and an agency's jurisdiction were insufficient unless the source identified the candidate county or a location within it. Prospective drought risk, general seasonal dryness, forecasts, watches, warnings, preparedness notices, and other statements about possible future conditions did not establish that the corresponding drought or extreme precipitation hazard had occurred during the applicable candidate window. Similarly, possible losses, exposed asset values, funding announcements, and eligibility for assistance did not establish a local impact. A single accepted webpage could support several assessments when separate passages documented the antecedent drought, the subsequent extreme precipitation hazard, a local consequence, the connection between that consequence and the hazard, or an explicit connection between the drought and the later extreme precipitation event.

\subsection{Assigning the Five Decisions}

The U.S. Drought Monitor uses D0 for abnormally dry conditions and D1–D4 for progressively more severe drought, ranging from moderate drought at D1 to exceptional drought at D4. The antecedent-drought assessment asked whether qualifying evidence documented drought in the candidate county during the antecedent drought window. A U.S. Drought Monitor county record supported the assessment when the relevant reporting period showed D1-or-higher conditions. A public webpage could also support the assessment when it explicitly documented drought, a drought-related water shortage, or drought-related restrictions in the candidate county during the candidate window and provided a verifiable supporting passage. The subsequent-extreme precipitation-event assessment asked whether an observed rain-, flood-, or precipitation-related event occurred in the candidate county during the extreme precipitation-event window. The impact assessment asked whether a documented local consequence was reported for the matched extreme precipitation event, such as property or crop damage, road disruption, power interruption, evacuation, debris flow, landslide, casualty, or rescue activity. The hazard impact assessment asked whether the source connected a documented consequence to rainfall, flooding, or another matched extreme precipitation hazard in the same event. The drought-to-extreme precipitation assessment required a source to explicitly connect the antecedent drought conditions to the subsequent rainfall, flooding, or extreme precipitation event. The chronological occurrence of drought followed by rainfall was not sufficient by itself.

Each assessment received a Yes, No, or Unresolved decision. A Yes decision required at least one eligible source that passed all applicable date, location, event, and content checks. Because the five questions were assessed separately, the same source could support more than one decision, while evidence relevant to one question did not automatically support another.

An Unresolved decision was assigned when no source supported Yes, but the available material did not permit a complete judgment. This included potentially relevant sources whose event, date, or location could not be matched conclusively, sources whose text was insufficient to evaluate, and cases in which the required review could not be completed. A No decision was assigned only after the planned retrieval and review for that assessment had been completed, no source met the requirements for Yes, and no unresolved source prevented a decision. A No decision therefore means that the completed review found no qualifying record under the stated evidence requirements; it does not establish that the event, impact, or relationship was absent in the real world.

For every candidate, the system retained the official-record identifiers, webpage queries and URLs, relevant source text, supporting quotations, extracted dates and locations, event-matching results, recorded retrieval limitations, and the rule producing each of the five final decisions. These retained materials allow each decision to be checked against the exact source information on which it was based.

\subsection{Human Validation of Retrieved Records}

To verify that the retrieved records actually documented the events detected from the observed data, we manually audited 40 random cases from the California candidate set. For each case, the audit examined the antecedent drought and the subsequent extreme precipitation event separately. A record counted as supporting an event only when its content matched the candidate county, fell within the relevant event window, and described the corresponding drought or extreme precipitation hazard rather than merely mentioning related terms.

When the pipeline had identified a supporting record, reviewers examined the cited record and its supporting passage or structured entry, checking the location, dates, and hazard type against the case. When the pipeline had not established support, reviewers searched public records for qualifying sources that retrieval may have missed. For these searches, the corresponding pipeline materials were withheld until the reviewer submitted the search result. This procedure tested both whether the records accepted by the pipeline matched the case and whether relevant records had been missed.

Both reviewers confirmed every drought or extreme precipitation-event record accepted by the pipeline and independently found the same two additional extreme precipitation-event records. Across all 40 audited cases, manual review matched the pipeline for all 40 drought assessments and 38 of 40 extreme precipitation-event assessments. This corresponds to agreement on 78 of 80 event assessments (97.5\%) and on both events in 38 of 40 cases (95.0\%). No event supported by the pipeline was rejected during manual review; both differences were extreme precipitation-event records missed during retrieval.

\section{Results}


\subsection{Record Support for CDEP Candidates}
\begin{figure}[t]
    \centering
    \includegraphics[width=0.95\linewidth]{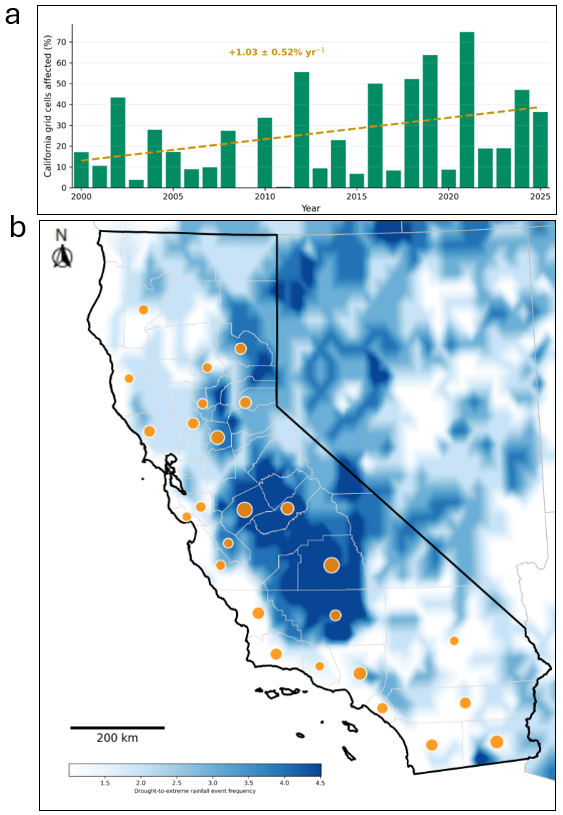}
    \caption{Historical expansion and recent spatial distribution of CDEP events in California. (a) Annual percentage of California grid cells affected by at least one CDEP event during 2000--2025; the dashed line shows the linear trend ($1.03 \pm 0.52$ percentage points per year). (b) CDEP event frequency during 2021--2025. Orange points indicate locations with documentary evidence accepted in the final review, with point size proportional to the number of accepted records at each location.}
    \label{fig:cdep_spatiotemporal_distribution}
\end{figure}

The proportion of California grid cells affected by at least one CDEP event increased during 2000--2025, with a linear trend of $1.03 \pm 0.52$ percentage points per year (Figure~\ref{fig:cdep_spatiotemporal_distribution}(a)). We focus the documentary review on 408 candidates identified during 2021--2025. These candidates were distributed across diverse regions of California (Figure~\ref{fig:cdep_spatiotemporal_distribution}(b), blue shade) and had a median drought-to-extreme-precipitation interval of approximately 35 days.

We applied the same retrieval and labeling procedure to all 408 California drought-to-extreme-precipitation candidates identified from the physical data. For each candidate, we evaluated the antecedent drought and the subsequent extreme-precipitation event separately using structured public records and public webpages. We counted a candidate as corroborated only when both components received a Yes judgment. At the candidate level, No means that at least one required component received a No judgment. Unresolved means that neither component received a No judgment, but at least one component remained unresolved.

\begin{table}[t]
\centering
\small
\caption{Record support for the 408 detected drought-to-extreme-precipitation candidates. A candidate is counted as corroborated only when records support both the antecedent drought and the subsequent extreme-precipitation event.}
\label{tab:candidate_corroboration}
\setlength{\tabcolsep}{4pt}
\renewcommand{\arraystretch}{1.1}
\begin{tabular}{@{}lccc@{}}
\toprule
Assessment & Yes & No & Unresolved \\
\midrule
Antecedent drought
& 297 (72.8\%)
& 111 (27.2\%)
& 0 (0.0\%) \\

\shortstack[l]{Subsequent extreme-\\precipitation event}
& 187 (45.8\%)
& 160 (39.2\%)
& 61 (15.0\%) \\

Both components
& 140 (34.3\%)
& 217 (53.2\%)
& 51 (12.5\%) \\
\bottomrule
\end{tabular}
\end{table}

Records supported the antecedent drought in 297 candidates (72.8\%) and the subsequent extreme-precipitation event in 187 candidates (45.8\%). Both components were supported in 140 candidates (34.3\%). These 140 cases form the set for which the retrieved records corroborated both parts of the detected drought-to-extreme-precipitation sequence. Support for only one component was not sufficient for a candidate to enter this set.

Among the remaining candidates, 217 had no qualifying support for at least one required component. Another 51 remained unresolved because the retrieved records did not support a firm extreme-precipitation event judgment for the relevant county and event window. We evaluated local impacts, hazard--impact connections, and explicit drought-to-extreme-precipitation connections separately. These results are examined below, with the complete Yes, No, and Unresolved distributions provided in the supplementary material. Accepted documentary evidence was geographically uneven and occurred at only a subset of detected-event locations (Figure~\ref{fig:cdep_spatiotemporal_distribution}(b); orange points, with size proportional to the number of accepted records).

\subsection{Event Characteristics Associated with Public-Web Corroboration}

Because the agent found public webpages for only a subset of the meteorologically detected candidates, we tested whether documentary corroboration varied systematically with measurable characteristics of the detected events. This analysis characterizes which candidates were more likely to appear in accessible public webpages under a common matching procedure.

We next compared candidates with and without at least one public
webpage that passed the date, location, event, and content checks
described above. We considered four characteristics of the detected
sequence: extreme-precipitation event duration, the interval between the end of the
drought and the start of the extreme-precipitation event, drought extremity, and
antecedent drought duration. Table~\ref{tab:webpage_characteristics}
reports the group summaries and odds ratios from a logistic regression
containing all four characteristics.

Extreme-precipitation event duration showed the clearest difference. Candidates with a
qualifying webpage had extreme-precipitation events lasting 3.11 days on average, compared
with 2.66 days for the other candidates. In the joint model, each
additional day of extreme-precipitation event duration was associated with 1.40 times
the odds of finding a qualifying webpage (95\% CI 1.04--1.88;
$p=0.025$). The estimate changed little when standard errors were clustered by date-defined extreme-precipitation event group and remained similar after candidates with recorded retrieval or coverage failures were excluded
(OR $=1.37$).

Candidates with qualifying webpages also had shorter drought-to-extreme-precipitation
intervals on average, 41.9 days compared with 49.1 days. The adjusted
odds ratio was 0.72 for each additional 30 days
(95\% CI 0.47--1.10). Mean drought extremity was 3.20 in the qualifying-webpage group and 2.62 in the other group; the adjusted odds ratio was 1.21 per unit (95\% CI 0.94--1.57).
Antecedent drought duration averaged 7.11 months in the
matching-webpage group and 6.14 months in the other group, with an
adjusted odds ratio of 1.02 per three months
(95\% CI 0.76--1.37). Among the four characteristics, extreme-precipitation event
duration showed the clearest relation to whether a qualifying webpage
was found.

This result shows that meteorologically detected candidates were not equally represented in the accessible public-web record; the observed difference may reflect event salience, reporting practices, source availability, or retrieval coverage.

\begin{table}[t]
\centering
\caption{Candidate characteristics by qualifying-webpage status.}
\label{tab:webpage_characteristics}

\footnotesize
\setlength{\tabcolsep}{1.2pt}
\renewcommand{\arraystretch}{1.12}

\begingroup
\footnotesize
\setlength{\tabcolsep}{2pt}
\renewcommand{\arraystretch}{1.12}

\begin{tabular}{
@{}
>{\raggedright\arraybackslash}m{0.29\columnwidth}
>{\centering\arraybackslash}m{0.18\columnwidth}
>{\centering\arraybackslash}m{0.20\columnwidth}
>{\centering\arraybackslash}m{0.28\columnwidth}
@{}
}
\toprule

\textbf{Characteristic}
&
\textbf{Qualifying webpage}
&
\textbf{No qualifying webpage}
&
\textbf{Adjusted OR (95\% CI)}
\\
\midrule

Extreme-precipitation duration (days)
&
\mbox{3.11}
&
\mbox{2.66}
&
\mbox{1.40 (1.04--1.88)}
\\

Transition interval (days)
&
\mbox{41.9}
&
\mbox{49.1}
&
\mbox{0.72 (0.47--1.10)}
\\

Drought extremity
&
\mbox{3.20 (2.57)}
&
\mbox{2.62 (2.53)}
&
\mbox{1.21 (0.94--1.57)}
\\

Antecedent drought duration (months)
&
\mbox{7.11}
&
\mbox{6.14}
&
\mbox{1.02 (0.76--1.37)}
\\

\bottomrule
\end{tabular}

\endgroup

\vspace{2pt}

\begin{minipage}{\columnwidth}
\scriptsize
\raggedright
\textit{Note.} Values are means except drought extremity, reported
as mean (median). Transition interval is measured from drought
termination to extreme-precipitation onset. Adjusted ORs are from
the joint four-predictor model and are scaled per 1 day, 30 days,
1 unit, and 3 months, respectively.
\end{minipage}

\end{table}

\subsection{Local Impacts of Matched Extreme-Precipitation Events}

The record review extended beyond confirming that an extreme-precipitation hazard
occurred. It identified local impacts for 78 candidates and hazard--impact connections for 76 candidates. Qualifying public webpages were found for 35 of the candidates with local impacts and 33 of the candidates with hazard--impact connections. These records connected the detected
extreme-precipitation events to concrete consequences in the affected county.

Nearly nine in ten qualifying webpage records (40 of 45) documented at least one local impact, and more than four in five (37 of 45) explicitly connected that impact to rainfall or flooding. The recorded
consequences included flooding, road and traffic disruption, power
outages, evacuations and emergency actions, debris flows and
landslides, property damage, and rescue or medical response. The
qualifying webpages therefore did more than note heavy rainfall or
flooding: they documented how the event affected residents, roads,
homes, businesses, and infrastructure.

This pattern appeared throughout the study period and across different
parts of California. Qualifying webpages were found in every year from
2021 through 2025, across 26 storm families and 26
counties, including both major urban counties and less-populated
inland and rural counties. A smaller set of sources explicitly connected the antecedent drought to the later extreme-precipitation event. Together, these records described local consequences of the matched extreme-precipitation events and, in six cases, explicitly connected the extreme-precipitation event to antecedent drought conditions.

\FloatBarrier
\section{Discussion and Conclusion}

We developed the CDEP Agent to determine whether drought-to-extreme-precipitation transitions detected from meteorological data also appeared in official records and public webpages. The agent connected each meteorological candidate to records of the antecedent drought, the subsequent extreme-precipitation event, and any documented local consequences, while retaining the supporting sources, quotations, dates, locations, and event-matching results. Among 408 candidates identified in California, 140 were supported on both the drought and extreme-precipitation components. Only six candidates had a source that explicitly connected the antecedent drought conditions to the subsequent rainfall or flooding. These findings answer the central question posed in this study: a drought-to-extreme-precipitation sequence can be clearly identified from meteorological data even when the two hazards are documented separately, or when the relationship between them is not documented at all.

The methodological contribution of CDEP Agent is not simply
source retrieval, but a candidate-specific evidence-review
process. Each candidate fixes the county and event windows
before retrieval; follow-up searches target only missing
evidence, while deterministic date, location, and same-event
checks constrain how retrieved text affects the final decisions.
By retaining source passages, matching results, and decision
rules, the system produces auditable assessments rather than
untraceable model judgments.

The results also show what different records contribute to the review. Extreme-precipitation event duration had the clearest association with the presence of a qualifying public webpage: each additional day of the extreme-precipitation event was associated with 1.40 times the odds of finding a matching page, and the estimate remained similar in the sensitivity analyses. The corresponding relationships for drought extremity, antecedent drought duration, and the drought-to-extreme-precipitation interval were less clear. When a qualifying webpage was available, however, it usually described more than the occurrence of heavy rainfall or flooding. Forty of the 45 qualifying webpage records documented at least one local consequence, and 37 explicitly connected that consequence to rainfall or flooding. These pages recorded effects including road disruption, power outages, evacuations, property damage, debris flows, and rescue activity. Structured records were therefore useful for confirming drought and extreme-precipitation hazards, while public webpages often supplied the local consequences needed to understand how a matched event affected residents and infrastructure.

The broader impact of this work is to make meteorologically detected compound events easier to examine in relation to documented hazards and local consequences. Climate scientists can apply the same evidence-review procedure to candidate sets produced by different drought thresholds, rainfall thresholds, or transition windows and compare which definitions identify events that also appear in official or public records. Because meteorological candidate construction is separate from evidence review, changing an event definition does not require rebuilding the entire review process. Social scientists, economists, and disaster researchers can use the retained sources, quotations, dates, and locations to study documented consequences of the same set of meteorologically identified events. The results also show a limitation of existing event records for compound-event research: droughts and extreme-precipitation events may both be recorded, while the connection between them remains absent. A review procedure that preserves both components and their relationship can therefore support future datasets designed specifically for sequential compound climate events.

The main limitation of the present study is its geographic and temporal scope. Extending the analysis from California to the entire United States would require substantially more web retrieval, processing, and human review across states with different climates, county structures, government records, and local reporting practices. Given the computational and review resources available for this study, we evaluated candidates in California during 2021--2025. The exact support rates and regression estimates reported here should therefore not be assumed to remain unchanged in other regions or over longer periods. We also evaluated one operational definition based on SPEI-3, a location-specific precipitation threshold, and a three-month transition window. Alternative definitions may produce different candidate sets and different documentary patterns. The separation between candidate construction and evidence review nevertheless allows the same procedure to be applied in future work to other thresholds, longer study periods, additional regions, and other types of compound climate events.

Meteorological detection identifies where a compound event may have occurred; it does not by itself show how that event was recorded or what local consequences were reported. By connecting each candidate to inspectable hazard and impact evidence, the CDEP Agent provides a repeatable way to examine that missing part of the event record.

\clearpage

\bibliography{reference_verified}

@article{rodell2023,
  author = {Rodell, M. and Li, B.},
  title = {Changing intensity of hydroclimatic extreme events revealed by {GRACE} and {GRACE}-{FO}},
  journal = {Nature Water},
  volume = {1},
  pages = {241--248},
  year = {2023},
  doi = {10.1038/s44221-023-00040-5}
}

@article{gu2023,
  author = {Gu, L. and Yin, J. and Gentine, P. and others},
  title = {Large anomalies in future extreme precipitation sensitivity driven by atmospheric dynamics},
  journal = {Nature Communications},
  volume = {14},
  pages = {3197},
  year = {2023},
  doi = {10.1038/s41467-023-39039-7}
}

@article{satoh2022,
  author = {Satoh, Y. and Yoshimura, K. and Pokhrel, Y. and others},
  title = {The timing of unprecedented hydrological drought under climate change},
  journal = {Nature Communications},
  volume = {13},
  pages = {3287},
  year = {2022},
  doi = {10.1038/s41467-022-30729-2}
}

@article{qing2023,
  author = {Qing, Y. and Wang, S. and Yang, Z.-L. and Gentine, P.},
  title = {Soil moisture-atmosphere feedbacks have triggered the shifts from drought to pluvial conditions since 1980},
  journal = {Communications Earth \& Environment},
  volume = {4},
  pages = {254},
  year = {2023},
  doi = {10.1038/s43247-023-00922-2}
}

@techreport{ipcc2021,
  author = {{IPCC}},
  title = {Climate Change 2021: The Physical Science Basis. Chapter 11: Weather and Climate Extreme Events in a Changing Climate},
  institution = {Working Group I contribution to the Sixth Assessment Report},
  year = {2021}
}

@article{zscheischler2025,
  author = {Zscheischler, J. and Raymond, C. and Chen, Y. and others},
  title = {Compound weather and climate events in 2024},
  journal = {Nature Reviews Earth \& Environment},
  volume = {6},
  number = {4},
  pages = {240--242},
  year = {2025},
  doi = {10.1038/s43017-025-00657-y}
}

@article{zscheischler2020,
  author = {Zscheischler, J. and Martius, O. and Westra, S. and others},
  title = {A typology of compound weather and climate events},
  journal = {Nature Reviews Earth \& Environment},
  volume = {1},
  pages = {333--347},
  year = {2020},
  doi = {10.1038/s43017-020-0060-z}
}

@article{leonard2014,
  author = {Leonard, M. and Westra, S. and Phatak, A. and others},
  title = {A compound event framework for understanding extreme impacts},
  journal = {WIREs Climate Change},
  volume = {5},
  number = {1},
  pages = {113--128},
  year = {2014},
  doi = {10.1002/wcc.252}
}

@article{jones2022,
  author = {Jones, R. L. and Guha-Sapir, D. and Tubeuf, S.},
  title = {Human and economic impacts of natural disasters: Can we trust the global data?},
  journal = {Scientific Data},
  volume = {9},
  pages = {572},
  year = {2022},
  doi = {10.1038/s41597-022-01667-x}
}

@article{rosvold2021,
  author = {Rosvold, E. L. and Buhaug, H.},
  title = {{GDIS}, a global dataset of geocoded disaster locations},
  journal = {Scientific Data},
  volume = {8},
  pages = {61},
  year = {2021},
  doi = {10.1038/s41597-021-00846-6}
}

@article{teber2025,
  author  = {Teber, Khalil and Weynants, M{\'e}lanie and
             Gans, Fabian and Mahecha, Miguel D.},
  title   = {Geo-Disasters: Geocoding climate-related events
             in the international disaster database {EM-DAT}},
  journal = {Big Earth Data},
  year    = {2026},
  volume  = {10},
  number  = {1},
  pages   = {303--318},
  doi     = {10.1080/20964471.2025.2576274}
}

@article{yin2025,
  author = {Yin, C. and Ting, M. and Kornhuber, K. and Horton, R. M. and Yang, Y. and Jiang, Y.},
  title = {{CETD}: A global compound events detection and visualisation toolbox and dataset},
  journal = {Scientific Data},
  volume = {12},
  pages = {356},
  year = {2025},
  doi = {10.1038/s41597-025-04530-x}
}

@article{vahedifard2017,
  author = {Vahedifard, F. and AghaKouchak, A. and Ragno, E. and Shahrokhabadi, S. and Mallakpour, I.},
  title = {Lessons from the {O}roville dam},
  journal = {Science},
  volume = {355},
  number = {6330},
  pages = {1139--1140},
  year = {2017},
  doi = {10.1126/science.aan0171}
}

@article{swain2018,
  author = {Swain, D. L. and Langenbrunner, B. and Neelin, J. D. and Hall, A.},
  title = {Increasing precipitation volatility in twenty-first-century {C}alifornia},
  journal = {Nature Climate Change},
  volume = {8},
  pages = {427--433},
  year = {2018},
  doi = {10.1038/s41558-018-0140-y}
}

@article{swain2025,
  author = {Swain, D. L. and Prein, A. F. and Abatzoglou, J. T. and others},
  title = {Hydroclimate volatility on a warming {E}arth},
  journal = {Nature Reviews Earth \& Environment},
  volume = {6},
  pages = {35--50},
  year = {2025},
  doi = {10.1038/s43017-024-00624-z}
}

@misc{skarlinski2024,
  author = {Skarlinski, M. D. and Cox, S. and Laurent, J. M. and Braza, J. D. and Hinks, M. M. and Hammerling, M. J. and Ponnapati, M. and Rodriques, S. G. and White, A. D.},
  title = {Language agents achieve superhuman synthesis of scientific knowledge},
  howpublished = {arXiv:2409.13740},
  year = {2024},
  doi = {10.48550/arXiv.2409.13740}
}

@article{asai2026,
  author = {Asai, A. and He, J. and Shao, R. and others},
  title = {Synthesizing scientific literature with retrieval-augmented language models},
  journal = {Nature},
  volume = {650},
  pages = {857--863},
  year = {2026},
  doi = {10.1038/s41586-025-10072-4}
}

@inproceedings{chen2025,
  author = {Chen, Z. and Chen, S. and Ning, Y. and others},
  title = {{ScienceAgentBench}: Toward rigorous assessment of language agents for data-driven scientific discovery},
  booktitle = {International Conference on Learning Representations (ICLR 2025)},
  year = {2025}
}

@article{xie2025,
  author = {Xie, Y. and Jiang, B. L. and Mallick, T. and Bergerson, J. and Hutchison, J. and Verner, D. and Branham, J. and Alexander, M. R. and Ross, R. and Feng, Y. and Levy, L.-A. and Su, W. and Taylor, C.},
  title = {{MARSHA}: Multi-Agent {RAG} System for Hazard Adaptation},
  journal = {npj Climate Action},
  volume = {4},
  pages = {70},
  year = {2025},
  doi = {10.1038/s44168-025-00254-1}
}

@article{li2026wikimpacts,
  author = {Li, N. and Thiery, W. and Zahra, S. and others},
  title = {{Wikimpacts} 1.0: A new global climate impact database based on automated information extraction from {Wikipedia}},
  journal = {Natural Hazards and Earth System Sciences},
  volume = {26},
  number = {6},
  pages = {2609--2636},
  year = {2026},
  doi = {10.5194/nhess-26-2609-2026}
}

@article{munozsabater2019,
  author = {Mu{\~n}oz Sabater, J.},
  title = {{ERA5}-Land hourly data from 1950 to present},
  journal = {Copernicus Climate Change Service Climate Data Store},
  year = {2019},
  doi = {10.24381/cds.e2161bac}
}

@article{vicenteserrano2010,
  author = {Vicente-Serrano, S. M. and Begu{\'e}r{\'i}a, S. and L{\'o}pez-Moreno, J. I.},
  title = {A multiscalar drought index sensitive to global warming: The Standardized Precipitation Evapotranspiration Index},
  journal = {Journal of Climate},
  volume = {23},
  pages = {1696--1718},
  year = {2010},
  doi = {10.1175/2009JCLI2909.1}
}

@article{vicenteserrano2020,
  author = {Vicente-Serrano, S. M. and Dom{\'i}nguez-Castro, F. and McVicar, T. R. and others},
  title = {Global characterization of hydrological and meteorological droughts under future climate change: The importance of timescales, vegetation--{CO2} feedbacks and changes to distribution functions},
  journal = {International Journal of Climatology},
  volume = {40},
  number = {5},
  pages = {2557--2567},
  year = {2020},
  doi = {10.1002/joc.6350}
}

@article{deng2024,
  author = {Deng, S. and Zhao, D. and Chen, Z. and others},
  title = {Global distribution and projected variations of compound drought-extreme precipitation events},
  journal = {Earth's Future},
  volume = {12},
  pages = {e2024EF004809},
  number = {7},
  year = {2024},
  doi = {10.1029/2024EF004809}
}

@article{li2012,
  author = {Li, K. and Wu, S. and Dai, E. and Xu, Z.},
  title = {Flood loss analysis and quantitative risk assessment in {C}hina},
  journal = {Natural Hazards},
  volume = {63},
  number = {2},
  pages = {737--760},
  year = {2012},
  doi = {10.1007/s11069-012-0180-y}
}

@article{stein2020,
  author = {Stein, L. and Pianosi, F. and Woods, R.},
  title = {Event-based classification for global study of river flood generating processes},
  journal = {Hydrological Processes},
  volume = {34},
  number = {7},
  pages = {1514--1529},
  year = {2020},
  doi = {10.1002/hyp.13678}
}

\clearpage
\section*{Supplementary Material}
\setcounter{table}{0}
\renewcommand{\thetable}{S\arabic{table}}

\begin{table*}[t] \centering \caption{Evidence required and evidence treated as insufficient for the five candidate assessments.} \label{tab:supp_evidence_requirements} \small \setlength{\tabcolsep}{4pt} \renewcommand{\arraystretch}{1.12} \begin{tabularx}{\textwidth}{ @{} >{\raggedright\arraybackslash}p{0.17\textwidth} >{\raggedright\arraybackslash}X >{\raggedright\arraybackslash}X @{} } \toprule \textbf{Assessment} & \textbf{Evidence required for \textit{Yes}} & \textbf{Evidence that does not support \textit{Yes}} \\ \midrule Antecedent drought & A U.S. Drought Monitor county record reported D1-or-higher conditions for the relevant reporting period, or a qualifying public webpage documented drought or drought-related dry conditions, water shortage, or restrictions in the candidate county during the antecedent drought window and provided a verifiable supporting quotation. & General references to dry weather; prospective drought risk; conditions outside the candidate drought window; statewide or regional descriptions without a county-level location match; or a drought-focused query without source-grounded evidence. \\ \addlinespace Subsequent extreme-precipitation event & A rain-, flood-, or precipitation-related event occurred during the candidate extreme-precipitation event window and in the candidate county. Evidence from a public webpage also required a verifiable quotation describing the event. & Forecasts, watches, warnings, or preparedness notices; publication dates without corresponding event dates; events outside the candidate window; or statewide or regional descriptions without a county-level location match. \\ \addlinespace Local impact & A documented consequence was reported for the matched extreme-precipitation event, such as property or crop damage, road disruption, power interruption, evacuation, debris flow, landslide, casualty, rescue activity, or another local impact. & Potential losses; exposed asset values; assistance eligibility; funding announcements; preparedness notices; actions that were only proposed or recommended; or statements that impacts were possible without documenting an observed consequence. \\ \addlinespace Hazard--impact connection & A consequence was recorded for the same NOAA event, or a public webpage explicitly connected a local impact to rainfall, flooding, or another matched extreme-precipitation hazard. & For public webpages, a hazard and an impact that appeared in the same document but were not explicitly connected; references to different events; or a connection inferred by combining separate sources. \\ \addlinespace Explicit drought-to-extreme-precipitation connection & A webpage explicitly connected the antecedent drought or dry conditions to the subsequent rainfall, flooding, or extreme-precipitation event. & A drought record followed by an extreme-precipitation event record; general climatic background; or a connection inferred by combining separate sources. \\ \bottomrule \end{tabularx} \end{table*}

\subsection{Retrieval Planning and Query Controls}

Each candidate entered the retrieval stage with a county, an antecedent drought period, a subsequent extreme-precipitation event window, and the corresponding physical-event information. The system first matched the candidate against the U.S. Drought Monitor and NOAA Storm Events and stored the resulting official records in the candidate record. It then conducted an initial public-web search using the candidate location, event period, and terms describing the relevant hazard or consequence.

After the initial pass, the agent examined the accepted evidence already available for the candidate and selected additional searches according to the information that remained unsupported. A drought-focused search was used when no qualifying source documented the antecedent drought. An extreme-precipitation hazard search was used when no qualifying source documented the subsequent extreme-precipitation event. An impact-focused search was used when the available sources did not document a local consequence or provided only limited information about the consequences of a matched extreme-precipitation event. Once both the antecedent drought and the subsequent extreme-precipitation event were supported, a linkage-focused search could be used to locate a source that explicitly connected the dry period to the later rainfall or flooding.

\begin{table*}[t] \centering \caption{Targeted follow-up searches selected from each candidate's current evidence record. The same selection policy and retrieval limits were applied to all candidates; only the search targets relevant to the remaining evidence gaps were activated.} \label{tab:supp_followup_actions} \small \setlength{\tabcolsep}{5pt} \renewcommand{\arraystretch}{1.12} \begin{tabularx}{\textwidth}{ @{} >{\raggedright\arraybackslash}p{0.23\textwidth} >{\raggedright\arraybackslash}p{0.20\textwidth} >{\raggedright\arraybackslash}X @{} } \toprule \textbf{Evidence state after initial retrieval} & \textbf{Search target} & \textbf{Information sought} \\ \midrule No qualifying source documented the antecedent drought & Drought hazard & Observed drought or drought-related dry conditions, water shortage, or restrictions in the candidate county during the antecedent drought window. \\ \addlinespace No qualifying source documented the subsequent extreme-precipitation event & Extreme-precipitation hazard & Observed heavy rainfall, flooding, or another compatible extreme-precipitation hazard in the candidate county during the extreme-precipitation event window. \\ \addlinespace A matched extreme-precipitation event was available, but its local consequences were absent or incompletely described & Impact & Property or crop damage, road disruption, power interruption, evacuation, debris flow, landslide, casualty, rescue activity, or another documented local consequence. \\ \addlinespace The antecedent drought and subsequent extreme-precipitation event were both supported, but no source explicitly connected them & Explicit drought-to-extreme-precipitation connection & A source statement connecting antecedent drought or dry conditions to the subsequent rainfall, flooding, or extreme-precipitation event. \\ \bottomrule \end{tabularx} \end{table*}

\begin{table*}[t]
\centering
\small
\caption{Complete decision distributions for the 408 drought-to-extreme-precipitation
candidates.}
\label{tab:complete_decision_distributions}
\begin{tabular}{lccc}
\toprule
Assessment & Yes & No & Unresolved \\
\midrule
Antecedent drought
& 297 (72.8\%)
& 111 (27.2\%)
& 0 (0.0\%) \\

Subsequent extreme-precipitation event
& 187 (45.8\%)
& 160 (39.2\%)
& 61 (15.0\%) \\

Both drought and extreme-precipitation components
& 140 (34.3\%)
& 217 (53.2\%)
& 51 (12.5\%) \\
\midrule
Local impact
& 78 (19.1\%)
& 208 (51.0\%)
& 122 (29.9\%) \\

Hazard--impact connection
& 76 (18.6\%)
& 214 (52.5\%)
& 118 (28.9\%) \\

Explicit drought-to-extreme-precipitation connection
& 6 (1.5\%)
& 336 (82.4\%)
& 66 (16.2\%) \\
\bottomrule
\end{tabular}
\end{table*}

\begin{table*}[t]
\centering
\caption{Joint logistic regression for the presence of at least one
matching public webpage. All four predictors were included in the
same model.}
\label{tab:supp_joint_webpage_model}
\small
\setlength{\tabcolsep}{7pt}
\renewcommand{\arraystretch}{1.08}
\begin{tabular}{@{}lccccc@{}}
\toprule
\textbf{Predictor} &
\textbf{Coefficient} &
\textbf{SE} &
\textbf{Adjusted OR} &
\textbf{95\% CI} &
\textbf{$p$-value} \\
\midrule
Extreme-precipitation event duration, per day
& 0.336
& 0.150
& 1.40
& 1.04--1.88
& 0.025 \\
Drought-to-extreme-precipitation interval, per 30 days
& $-0.331$
& 0.217
& 0.72
& 0.47--1.10
& 0.127 \\
Drought extremity, per unit
& 0.191
& 0.131
& 1.21
& 0.94--1.57
& 0.144 \\
Antecedent drought duration, per 3 months
& 0.022
& 0.150
& 1.02
& 0.76--1.37
& 0.884 \\
\bottomrule
\end{tabular}
\end{table*}

The query-generation model received the candidate location, relevant dates, current evidence target, and the evidence already accepted for the case. Every generated query was checked before execution. A query had to retain the candidate county or another location already grounded in the case, include the relevant event period, and contain terms appropriate for the evidence being sought. Queries were rejected if they were too broad to identify the candidate event, duplicated or closely repeated an earlier query, or introduced storm names, roads, damage amounts, casualties, declarations, or other specific details that were not present in the candidate information or accepted evidence. URLs were normalized and deduplicated so that a page already considered in the initial pass was not opened again during follow-up retrieval.

The retrieval budget was fixed across candidates to keep the procedure comparable. The initial stage used at most one query and opened at most two webpages. The follow-up stage used at most four additional queries and opened at most three additional webpages. At most one follow-up query could target the antecedent drought and at most one could target the subsequent extreme-precipitation event. Impact-focused and linkage-focused searches retained maximums of three and one queries, respectively, but all search types shared the same four-query total. When both hazard-focused searches were activated, at most two follow-up queries remained for impact and linkage searches combined. Thus, each candidate used no more than five public-web queries and five opened webpages.

Search-result snippets and provider-generated summaries were used only to identify possible sources. They did not enter the evidence record. A webpage became eligible for review only after the underlying page text had been retrieved and stored. Pages that could not be retrieved or did not contain enough text for review were recorded in the candidate trace rather than treated as evidence that the event or impact was absent.

For each candidate, the running record stored the official evidence, generated queries, retrieved URLs, page-level review results, accepted and unresolved evidence, previously visited pages, and the remaining retrieval budget. This record allowed the follow-up search to depend on evidence already obtained for that candidate and preserved the actions leading to the final assessments.

\subsection{Page-Level Review and Structured Evidence}

Each retrieved webpage was reviewed separately together with the corresponding candidate information. The page-review model first determined whether the page contained no relevant record, one clearly identifiable drought period or extreme-precipitation event that could be matched to the candidate, or several periods or events that could not be separated reliably. Only a clearly matched record could contribute supporting evidence.

For a potentially matching record, the model returned a structured record containing the assessment being reviewed, the reported dates and locations, the observed drought or extreme-precipitation conditions relevant to that assessment, local impacts when applicable, stated relationships, and the passages supporting those fields.

The extracted record then passed explicit candidate-matching checks. The reported dates had to overlap the candidate window corresponding to the assessment under review: the antecedent drought window for drought evidence and the subsequent extreme-precipitation event window for extreme-precipitation event evidence. The source had to identify the candidate county or a clearly located place within that county; statewide descriptions, broad regional language, and an agency's jurisdiction were not sufficient by themselves. The page also had to describe an observed hazard rather than only a forecast, warning, preparedness notice, or statement of seasonal risk.

The same-event check required the hazard, consequence, and stated relationship to refer to one identifiable event. Information from different storms was not combined. When several events appeared on the same page but could not be separated reliably, the page was retained for review but did not support a positive decision. A documented consequence and an extreme-precipitation hazard appearing in the same page were also insufficient for the hazard--impact assessment unless the source explicitly connected them.

The language model therefore handled source-level interpretation, including event identification, date and location extraction, identification of observed hazards and consequences, and recognition of relationships explicitly stated in the source. Deterministic checks enforced date overlap, county matching, same-event consistency, quotation support, and the evidence requirements shown in Table~\ref{tab:supp_evidence_requirements}. This division allowed webpage prose to be interpreted while preventing an extracted field from supporting a decision unless it satisfied the candidate-specific checks.

The query-generation and webpage-review stages used GPT-5.5-2026-04-23. Query generation used temperature 0 and a maximum output length of 2,048 tokens. Page review used high reasoning effort and a maximum output length of 12,000 tokens. Both stages returned structured outputs. An unusable model response did not contribute evidence and was recorded as an incomplete review.

\subsection{Assessment Aggregation and Coverage Handling}

The accepted official and webpage records were aggregated separately for antecedent drought, subsequent extreme-precipitation event, local impact, hazard--impact connection, and explicit drought-to-extreme-precipitation connection. Each source retained its source identifier, relevant text or structured fields, extracted dates and locations, supporting passages, and matching results. A webpage could support more than one assessment when separate passages independently established the antecedent drought, the subsequent extreme-precipitation hazard, a local consequence, the connection between the consequence and the hazard, or an explicit drought-to-extreme-precipitation relationship.

Webpage evidence for antecedent drought and the subsequent extreme-precipitation event was kept separate. Evidence accepted or left uncertain for the drought assessment affected only the drought decision, and evidence accepted or left uncertain for the extreme-precipitation event assessment affected only the extreme-precipitation event decision. A webpage could support more than one assessment only when separate source-grounded passages independently met the requirements for each assessment; the search target alone did not count as evidence.

A Yes decision was assigned when at least one eligible source satisfied all requirements for the corresponding assessment. Once qualifying evidence established Yes, uncertainty in another retrieved source did not remove that supporting evidence. The candidate record nevertheless retained the additional source and its review status.

When no source supported Yes, the system distinguished a completed negative review from a review that could not be completed. A No decision required the relevant official and webpage checks to be complete, with no qualifying source, no potentially relevant source awaiting resolution, and no retrieval or processing failure that prevented evaluation. The scope of this decision was limited to the completed retrieval and review procedure.

An Unresolved decision was assigned when no source supported Yes but the available evidence did not permit a complete judgment. This included a potentially relevant source with an uncertain event match, insufficient page text, several inseparable events, an unavailable required record, or a retrieval or processing failure. An inaccessible or unreadable page could therefore not be used to produce No.

The final candidate record stored the five decisions together with the supporting and contextual source identifiers, the strongest date and location matches, coverage information, recorded review limitations, and the rule that produced each decision. These records were used to generate the decision distributions reported in Table~\ref{tab:complete_decision_distributions} and permit each result to be traced back to its source material.

\subsection{Representative Retrieval and Review Trace}

To illustrate how the stages operated together, consider the Sonoma County candidate with an antecedent drought extending from August 2020 through September 30, 2021, followed by an extreme-precipitation event window from October 24 to October 26, 2021. The U.S. Drought Monitor record supported the antecedent-drought assessment, while the initial structured records did not provide accepted evidence for the subsequent extreme-precipitation event or its local impacts.

The initial public-web query searched for flooding and emergency-management information in Sonoma County during the relevant period. The initially retrieved materials did not contain a sufficiently matched event. The agent therefore selected extreme-precipitation hazard and impact-focused follow-up searches while preserving the candidate location and event dates.

A subsequently retrieved local report described atmospheric-river rainfall and flooding in Sonoma County and Santa Rosa on October 23--24, 2021. The page reported water rescues, flooding of buildings and roads, and evacuation of residents. It explicitly connected the rainfall and overflowing waterways to these consequences. The same report described the event in the context of a drought-stricken year and stated that the downpour replenished waterways depleted during the drought.

The page-review model extracted the event dates, Sonoma County locations, observed rainfall and flooding, local consequences, hazard--impact statements, drought-to-extreme-precipitation statement, and the supporting passages. The date, county, and same-event checks accepted the event as a match to the candidate. Together with the U.S. Drought Monitor record, the accepted evidence supported \textit{Yes} decisions for antecedent drought, subsequent extreme-precipitation event, local impact, hazard--impact connection, and Explicit drought-to-extreme-precipitation connection. Other retrieved pages that did not pass the event-matching requirements remained in the candidate trace but did not contribute to these positive decisions.

\subsection{Complete Decision Distributions}

Table~\ref{tab:complete_decision_distributions} reports the complete
Yes, No, and Unresolved distributions for all five assessments. It
also reports whether the retrieved records supported both the
antecedent drought and the subsequent extreme-precipitation event. Percentages use all
408 candidates as the denominator.

For the both-components result, Yes required both the antecedent
drought and the subsequent extreme-precipitation event to receive Yes. No indicates
that at least one of the two components received No. Unresolved
indicates that neither component received No but at least one
remained Unresolved. Local impact, hazard--impact connection,
and explicit drought-to-extreme-precipitation connection were assessed separately and did not
affect the both-components result.

\subsection{Regression Specification and Sensitivity Analyses}

We modeled whether each candidate had at least one public webpage
that passed the date, location, same-event, and content requirements
described above. The analysis included all 408 candidates, and no
candidate was excluded because of missing outcome or predictor
values. We fitted an unpenalized logistic regression with an
intercept and included four candidate characteristics in the same
model: drought extremity, antecedent drought duration, extreme-precipitation event
duration, and the drought-to-extreme-precipitation interval.

Extreme-precipitation event duration was measured as the inclusive number of calendar
days in the candidate rainfall window. The drought-to-extreme-precipitation interval
was measured from the final day of the drought-ending month to the
start of the extreme-precipitation event window. Antecedent drought duration was the
inclusive number of months between the beginning and end of the
drought episode. Drought extremity was defined as the absolute value
of the minimum SPEI-3 value during the antecedent drought episode, so
larger values indicated more extreme drought. Odds ratios are
reported for a one-day increase in extreme-precipitation event duration, a 30-day
increase in the drought-to-extreme-precipitation interval, a one-unit increase in
drought extremity, and a three-month increase in antecedent drought
duration.

The confidence intervals and $p$-values in
Table~\ref{tab:supp_joint_webpage_model} were calculated using
model-based standard errors and normal-Wald inference. We conducted
two sensitivity analyses for the association between extreme-precipitation event
duration and the presence of a matching public webpage.

First, we recalculated the uncertainty using
finite-sample-corrected cluster-robust standard errors. Candidates
with identical extreme-precipitation event start and end dates were assigned to the
same group, producing 131 date-defined storm groups. The adjusted
odds ratio for extreme-precipitation event duration remained 1.40, with a clustered
95\% confidence interval of 1.04--1.88 and $p=0.026$.

Second, we refitted the same four-variable model after excluding 68
candidates with recorded extreme-precipitation event retrieval, source, access, or
coverage failures. The remaining analysis included 340 candidates.
The adjusted odds ratio for extreme-precipitation event duration was 1.37
(95\% CI 1.02--1.85; $p=0.038$). The direction and magnitude of the
association were similar in both sensitivity analyses.

\end{document}